\documentclass[10pt,onecolumn,letterpaper]{article}

\usepackage{arxiv} 
\usepackage[utf8]{inputenc} 
\usepackage[T1]{fontenc}    
\usepackage{url}            
\usepackage{booktabs}       
\usepackage{amsfonts}       
\usepackage{nicefrac}       
\usepackage{microtype}      
\usepackage{xcolor}         

\usepackage{graphicx}
\usepackage{amsmath}
\usepackage{amssymb}
\usepackage{booktabs}
\usepackage[utf8]{inputenc} 
\usepackage[T1]{fontenc}    
\usepackage{url}            
\usepackage{amsfonts}       
\usepackage{nicefrac}       
\usepackage{microtype}      
\usepackage{bbding}
\usepackage{pifont}
\usepackage{bbm}
\usepackage{bm}
\usepackage{soul}
\usepackage{multirow}
\usepackage{framed}
\usepackage{amsthm}
\usepackage{makecell}
\usepackage{verbatim}

\usepackage{amsmath}
\usepackage{amssymb}
\usepackage{mathtools}
\usepackage{amsthm}
\allowdisplaybreaks[4]

\usepackage{wrapfig}
\usepackage{caption}

\usepackage{amsmath}
\usepackage{amssymb}
\usepackage{mathtools}
\usepackage{amsthm}

\usepackage[pagebackref,breaklinks,colorlinks]{hyperref}

\theoremstyle{plain}
\newtheorem{theorem}{Theorem}

\theoremstyle{definition}

\theoremstyle{remark}
\newtheorem{remark}[theorem]{Remark}

\usepackage[textsize=tiny]{todonotes}

\usepackage[capitalize,noabbrev]{cleveref}

\usepackage[capitalize]{cleveref}
\crefname{section}{Sec.}{Secs.}
\Crefname{section}{Section}{Sections}
\Crefname{table}{Table}{Tables}
\crefname{table}{Tab.}{Tabs.}

\begin{document}

\title{Can the Inference Logic of Large Language Models be Disentangled into Symbolic Concepts?}

\author{
\textbf{Wen Shen}\quad\textbf{Lei Cheng}\quad\textbf{Yuxiao Yang}\quad\textbf{Mingjie Li}\quad\textbf{Quanshi Zhang}\thanks{Quanshi Zhang is the corresponding author. He is with the Department of Computer Science and Engineering, the John Hopcroft Center at Shanghai Jiao Tong University, China. \texttt{<zqs1022@sjtu.edu.cn>}}\\[2pt]
Shanghai Jiao Tong University\\
}
\maketitle

\begin{abstract}
In this paper, we explain the inference logic of large language models (LLMs) as a set of symbolic concepts. Many recent studies \cite{dengdiscovering,ren2023defining,ren2022can} have discovered that traditional DNNs usually encode sparse symbolic concepts. However, because an LLM has much more parameters than traditional DNNs, whether the LLM also encodes sparse symbolic concepts is still an open problem. Therefore, in this paper, we propose to disentangle the inference score of LLMs for dialogue tasks into a small number of symbolic concepts. We verify that we can use those sparse concepts to well estimate all inference scores of the LLM on all arbitrarily masking states of the input sentence. We also evaluate the transferability of concepts encoded by an LLM and verify that symbolic concepts usually exhibit high transferability across similar input sentences. More crucially, those symbolic concepts can be used to explain the exact reasons accountable for the LLM's prediction errors.
\end{abstract}

\section{Introduction}\label{sec:intro}

In recent years, large language models (LLMs), \emph{e.g.}, ChatGPT \cite{chatgpt} and GPT-3 \cite{brown2020language}, have exhibited remarkable performance. Although LLMs have been used to guide people in important tasks, such as writing papers, analyzing reports, and consulting information, the inference logic of LLMs still cannot be fully understood by people.

Therefore, generating an \emph{accountable} explanation for the LLM's inference score is critical for high-stake decisions. Generally speaking, an accountable explanation needs to satisfy three requirements.
\\
$\bullet$ \textbf{Countable inference patterns.}
Transforming the implicit inference logic of an LLM into explicit and countable inference patterns is crucial to explain the LLM. Specifically, it is found that the inference score of a deep neural network (DNN) on the input sentence can be disentangled into the effects of a small number of inference patterns (\emph{e.g.}, certain phrases) \cite{ren2023defining,li2023does}. These inference patterns can be considered as concepts memorized by the LLM, and they are supposed to universally explain inference scores on various sentences.
\\
$\bullet$ \textbf{Verified in practice.}
When the inference logic of an LLM is explained as a set of inference patterns, the faithfulness or accuracy of the explanation must be both guaranteed by theory and verified in experiments.
\\
$\bullet$ \textbf{Accountable for errors.}
When an LLM makes an incorrect prediction, it is important to clarify the exact reasons accountable for the error with a clear mathematical connection between the reason and the incorrect prediction, rather than provide a specious explanation.

\textbf{Definition of concepts in DNNs.} Considering the above three requirements, Ren et al. \cite{ren2023defining} have tried to explain the inference logic of a neural network as the detection of salient interactions between input variables. Specifically, they find that the network's inference score can be well mimicked by a small number of salient interactions between input variables. Given an input sentence with $n$ words indexed by $N = \{1, 2, ..., n\}$, the DNN does not directly use a single input word for inference. Instead, the DNN encodes the interactions between different words as inference patterns (or concepts) to generate an inference score $v$. For example, in the classification task, given the input sentence ``\emph{he is a green hand},'' the interaction between the words $S=\{\textit{green},\textit{hand}\} \subseteq N$ contributes a specific interaction effect $I(S)$, which pushes the network's inference score towards inferring ``\emph{he}'' to the meaning ``\emph{beginner}.'' Masking any words in $S$ will remove the interaction effect, \emph{i.e.}, making $I(S)=0$. 

It has been discovered that among all $2^n$ interactions, \emph{a DNN usually only encodes a small number of salient interactions, and all other interactions have ignorable effects $I(S)\approx 0$} \cite{ren2023defining,li2023does}. In this way, we can consider these salient interactions as \textbf{concepts} (\emph{i.e.}, countable inference patterns), and consider the rest interactions with ignorable effects as noisy patterns. Inspired by this, the inference logic of a DNN can be explained as the detection of a small number of concepts. In other words, \textbf{we can use such a few salient interaction concepts to well estimate inference scores of the DNN on an exponential number of all randomly masked sentences.}
More crucially, when a DNN makes incorrect predictions, the clear disentanglement of interaction concepts from the DNN allows us to identify the exact interaction concepts encoded by the DNN that are \textbf{accountable for errors}. In addition, those concepts have also exhibited high transferability across different samples and different DNNs, and demonstrated a considerable discrimination power in classification tasks \cite{li2023does}.

\textbf{Explaining LLMs using concepts.} However, all above findings of using sparse concepts to mimic the logic of a traditional DNN are obtained based on empirical observations in experiments without solid proof. On the other hand, the LLM usually has much more parameters than traditional DNNs, and the LLM is trained on much more data. Therefore, whether an LLM can be explained as the detection of a small number of concepts is still unknown.

Therefore, in this paper, we aim to disentangle the inference score of LLMs for dialogue into a small number of salient interaction concepts. Given an input sentence, \emph{e.g.}, a question or an unfinished sentence, the LLM sequentially generates words to answer the question or to complete the sentence. Considering the generation of the next word, the LLM takes the previous $n$ words as the input, and then it estimates the probability distribution of generating the $(n+1)$-th word. The LLM samples the word with the highest probability as the target word $y^{\text{truth}}$. Therefore, we set $v(\boldsymbol{x})=\log p(y=y^{\text{truth}} | \boldsymbol{x}) / (1-p(y=y^{\text{truth}} | \boldsymbol{x}))\in\mathbb{R}$ as the inference score of the LLM, where $p(y=y^{\text{truth}} | \boldsymbol{x})$ denotes the probalibity of generating the target $(n+1)$-th word $y^{\text{truth}}$. Specifically, we obtain the following three conclusions.
\\
1. We have verified that given an input sentence, we can use a small number of concepts to accurately estimate all inference scores of the LLM on all arbitrarily masking states of the input sentence. Given a well-trained LLM for dialogue, we quantify the interactions between different words in the input sentence, and verify that the output score of generating a specific word\footnote{Here, we focus on the language generation task that the LLM generates words sequentially.} can be well estimated using a few interaction concepts. Moreover, given an input sentence where some words are randomly masked, we can still use those interaction concepts to well estimate the output score of generating the next word.
\\
2. We have verified that interaction concepts encoded by the LLM exhibit high transferability across similar input sentences. In other words, given similar input sentences, the LLM usually uses similar interaction concepts to generate the same word. 
\\
3. We have verified that the disentangled interaction concepts can help people identify the exact reasons for the prediction error made by the LLM. When the LLM generates a wrong word that violated the facts, we find out those interaction concepts that are accountable for the wrong word.

\section{Revisiting literature in game-theoretic interactions to explain DNNs}

This paper is a typical achievement in the explanation system of game-theoretic interactions. Recently, our research group has built up a theoretic system of using game-theoretic interactions to explain black-box AI models in the following three aspects.

$\bullet$ First, we use game-theoretic interactions to \textbf{explain the knowledge/concepts encoded by a DNN}. For a long time, defining interactions between input variables of a DNN in game theory presents a new direction to explain a DNN \cite{grabisch1999axiomatic,sundararajan2020shapley}. Our research group has defined the multi-variate interaction \cite{zhang2021building,zhang2021interpreting} and the multi-order interaction \cite{zhanginterpreting2021} to represent interactions of different complexities. We also defined the optimal baseline value for representing the absence of the variable, which ensures the trustworthiness of masking-based attribution methods \cite{ren2022can}.

\emph{Extending from interactions to concepts.} In particular, recent studies of our group \cite{ren2023defining,li2023does} have discovered that a DNN usually encodes very sparse interactions when we used the Harsanyi interactions between input variables of the DNN. These studies actually suggested that game-theoretic interactions can be considered as the concepts encoded by a DNN. This conclusion was obtained based on the following three discoveries. (1) It has been found that a well-trained DNN usually only encoded a small number of salient interactions, and each interaction made a certain effect on the DNN's inference score. (2) It has been discovered that we could use such a small number of salient interactions to well estimate inference scores of all arbitrarily masked input samples. (3) It has been discovered that those salient interactions usually exhibited strong transferability across different samples and across different DNNs, and these salient interactions also exhibited strong discrimination power. 

The above three findings can be considered as the foundation to explain the inference logic of a DNN as the detection of salient interactions between input variables. For example, our research group used the multi-variate interaction to discover prototypical visual concepts encoded by a DNN \cite{cheng2021hypothesis}, and used the multi-order interaction to analyze the specific signal-processing behaviors of a DNN in encoding shapes and textures \cite{cheng2021game}. 

$\bullet$ Second, we use game-theoretic interactions to \textbf{investigate the representation
power of a DNN}. For example, we used the multi-order interaction to explain the generalization power of a DNN brought by the dropout operation \cite{zhang2021interpreting}, and to explain the adversarial
robustness and adversarial transferability of a DNN \cite{wang2021interpreting,wangunified,ren2023defining}.
Deng et al. \cite{dengdiscovering} discovered that a DNN had difficulty in representing interactions between intermediate number of input variables. Zhang et al. \cite{zhou2023concept} found that complex interactions (\emph{i.e.}, interactions between a large number of input variables) were more likely to be over-fitted. Furthermore, Ren et al. \cite{ren2023bayesian} discovered that a Bayesian neural network (BNN) was less likely to encode complex interactions, which avoided the problem of overfitting.

$\bullet$ Third, we use game-theoretic interactions to \textbf{analyze the common mechanism shared by
many empirical findings}. Deng et al. \cite{deng2022understanding} proved that the core mechanisms of fourteen attribution methods could be reformulated as a reallocation of interactions, which enables fair comparisons between attribution methods that were built upon different heuristics. Zhang et al. \cite{zhang2022proving} proved the common mechanism shared by twelve previous transferability-boosting methods, that is, these methods all reduced interactions between regional adversarial perturbations.

\section{Explaining the inference logic of large language models}

\subsection{Preliminaries: explaining inference logic with interaction concepts}

Recently, a series of studies \cite{ren2023defining,li2023does} have found that a DNN's output score can be decomposed into numerical effects of various interactions between different input variables, \emph{e.g.}, the interaction between different regions in an input image and the interaction between different words in an input sentence. Specifically, given a pre-trained DNN $v:\mathbb{R}^n \rightarrow \mathbb{R}$, let $\boldsymbol{x}$ denote the input sample, which has $n$ variables indexed by $N = \{1, 2, ..., n\}$. Without loss of generality, let us just focus on a scalar output of the DNN $v(\boldsymbol{x})\in\mathbb{R}$, \emph{e.g.}, a dimension of the output vector.
The DNN usually encodes interactions between different variables for inference. Each interaction has a numerical contribution to the network output $v(\boldsymbol{x})$. To this end, Ren et al. \cite{ren2023defining} have defined the following metric to measure the interaction between input variables in $S \subseteq N$, which has been encoded by the DNN.
\begin{equation}\label{eq:IS}
	\begin{aligned}
		I(S|\boldsymbol{x}) \triangleq \sum\nolimits_{T\subseteq S} (-1)^{\lvert S\rvert - \lvert T \rvert} \cdot v(\boldsymbol{x}_T).
	\end{aligned} 
\end{equation}
where $v(\boldsymbol{x}_T)$ denotes the network output when we mask variables in $N\setminus T$ and keep variables in $T$ unchanged.

$\bullet$ \textbf{Understanding the interaction in the LLM for the dialogue task.} Given a trained LLM for dialogue, let us understand the interaction defined in Equation (\ref{eq:IS}).
The LLM sequentially generates words to complete the input sentence. 
For example, given the input sentence ``\emph{he is a green hand in painting means that}'', the ChatGPT \cite{chatgpt} completes the input sentence as follows, ``\emph{he is a green hand in painting means that he is a beginner or novice in painting}.'' In this case, the LLM generates words ``\emph{he},'' ``\emph{is},'' ``\emph{a},'' ``\emph{beginner},'' etc., \textbf{one by one}. Thus, when the LLM has already generated $n$ words, denoted by $\boldsymbol{x}=[x_1,x_2,\ldots,x_n]^{\top}$, we analyze the probability of the LLM generating the $(n+1)$-th word. For example, when the LLM has generated ``\emph{he is a green hand in painting means that he is a},'' and then we insert this sentence as the input $\boldsymbol{x}$ to the LLM. Then, we analyze the probability of the LLM generating the target word $y^{\text{truth}}=$ [\emph{beginner}].
The inference score of the LLM generating the target word $y^{\text{truth}}$ can be given as follows. Note that for different tasks, $v(\boldsymbol{x})$ can be applied with different settings.
\begin{equation}
	v(\boldsymbol{x})=\log \frac{p(y = y^{\text{truth}} |\boldsymbol{x})}{1-p(y= y^{\text{truth}} |\boldsymbol{x})}
\end{equation}
Then, accordingly, $v(\boldsymbol{x}_T)$ in Equation (\ref{eq:IS}) corresponds to the inference score when we mask the words in $N\setminus T$ in the input sentence. An LLM typically uses the padding token (\emph{e.g.}, the ``PAD'' token used by the OPT-1.3b model \cite{zhang2022opt}) as a placeholder, which does not contain any semantic meanings. Therefore, we generate the masked sentense $\boldsymbol{x}_T$ by keeping words in $T$ unchanged and using the padding token to mask the words in $N\setminus T$.

In this way, the interaction $I(S|\boldsymbol{x})$ between a specific set of words in $S$ can be understood as the effect of this interaction on the LLM generating the target word. $I(S|\boldsymbol{x})>0$ indicates that the interaction between words in $S$ has positive effects on generating the target word. $I(S|\boldsymbol{x})<0$ indicates that the interaction prevents the LLM from generating the target word. $I(S|\boldsymbol{x})\approx 0$ indicates that the interaction almost has no effects on generating the target word.

More precisely, the interaction $I(S|\boldsymbol{x})$ represents an AND relationship between words in $S$.
In the above example ``he is a green hand in painting means that he is a beginner,'' let us consider two words of $S=\{\textit{green},\textit{hand}\}$ have strong interaction effect. Then, only when both words of ``\emph{green}'' and ``\emph{hand}'' co-appear in the input sentence, the interaction between $S=$ \{\emph{green},\emph{hand}\} is activated and contributes a numerical effect $I(S|\boldsymbol{x})$ to push the LLM to generate the word ``\emph{beginner}.'' Otherwise, if any word in $S$ is masked, then the LLM's output will not reveice the effect of this interaction, \emph{i.e.}, making $I(S|\boldsymbol{x})=0$.

$\bullet$ \textbf{Sparse-yet-universal matching.} Given a DNN $v$ and an input sample $\boldsymbol{x}$  with $n$ variables, we can generate $2^n$ different masked samples $\{\boldsymbol{x}_S|S\subseteq N\}$. Theorem~\ref{th:faith} shows that we can use interactions defined in Equation~(\ref{eq:IS}) to universally mimic the DNN's outputs on all the $2^n$ masked samples. For simplicity, we use $I_T$ to represent $I(T|\boldsymbol{x})$.
\begin{theorem}\label{th:faith}
	(\textbf{Faithfulness}, proven by Ren et al. \cite{ren2023defining}) 
	The DNN's outputs on all masked samples $\{\boldsymbol{x}_S|S\subseteq N\}$ can be universally mimicked as the sum of the triggered interaction effects, i.e., $\forall S\subseteq N, v(\boldsymbol{x}_S) = \sum\nolimits_{T\subseteq S} I_T$.
\end{theorem}
Furthermore, Ren et al. \cite{li2023does} have discovered that well-trained DNNs usually encode sparse interactions (see Remark~\ref{re:sparsity}). In other words, only a few interactions have salient influence on the DNN's outputs, \emph{i.e.}, the $\lvert I(S|\boldsymbol{x})\rvert$ values of these interactions are relatively large. In contrast, all other interactions have ignorable effects on the DNN's outputs, \emph{i.e.}, $I(S|\boldsymbol{x})\approx 0$, and can be considered as noisy patterns. Therefore, \textbf{we can use a few salient interactions with non-ignorable effects to well estimate the DNN's outputs on all $2^n$ different masked samples}, as follows, which theoretically guarantees the faithfulness of the explanation.
\begin{remark}\label{re:sparsity}
	\emph{(\textbf{Sparsity})
	The DNN's outputs on all masked samples $\{\boldsymbol{x}_S|S\subseteq N\}$ can be well estimated by a few salient interactions in $\Omega_{\text{salient}}$, subject to $\lvert \Omega_{\text{salient}} \rvert \ll 2^n$, i.e., $\forall S\subseteq N, v(\boldsymbol{x}_S) \approx \sum\nolimits_{T\in \Omega_{\text{salient}},T\subseteq S} I_T$.}
\end{remark}

According to Theorem~\ref{th:faith} and Remark~\ref{re:sparsity}, we can consider those \textbf{sparse salient interactions as symbolic concepts encoded by
the DNN.} For example, the salient interaction between two words in $S=\{\textit{green},\textit{hand}\}$ can be considered as a concept, which has the semantic meaning of ``\emph{beginner}.''

$\bullet$ \textbf{Transferability and discrimination power of symbolic concepts encoded by DNNs.} Li et al. \cite{li2023does} have discovered that many salient interactions extracted from a sample can also be found as salient interactions in another sample in the same category, \emph{i.e.}, salient interactions have considerable transferability across different samples in the same category. Besides, those sparse salient interactions also exhibit certain discrimination power in the classification task. That is, the same interaction usually has consistently positive effects or consistently negative effects on the classification of different samples in the same category.

\subsection{Exploring the concepts represented by the LLM}

In this subsection, we conducted experiments to diagnose feature representations encoded by the LLM.

$\bullet$ \textbf{Experiment 1: can the inference score of the LLM be disentangled into symbolic concepts?}
Previous studies have discovered that traditional DNNs usually encode sparse symbolic concepts. However, the LLM has much more parameters than traditional DNNs. There are two conflicting understanding of the LLM. 1. The LLM encodes much more complex concepts than traditional DNNs due to its large parameter number, or 2. alternatively, the LLM encodes more sparse concepts, because the LLM is usually more sophisticatedly trained and thus learns much clearer features.

\begin{figure}[tbp]
	\centering
	\includegraphics[width=0.98\linewidth]{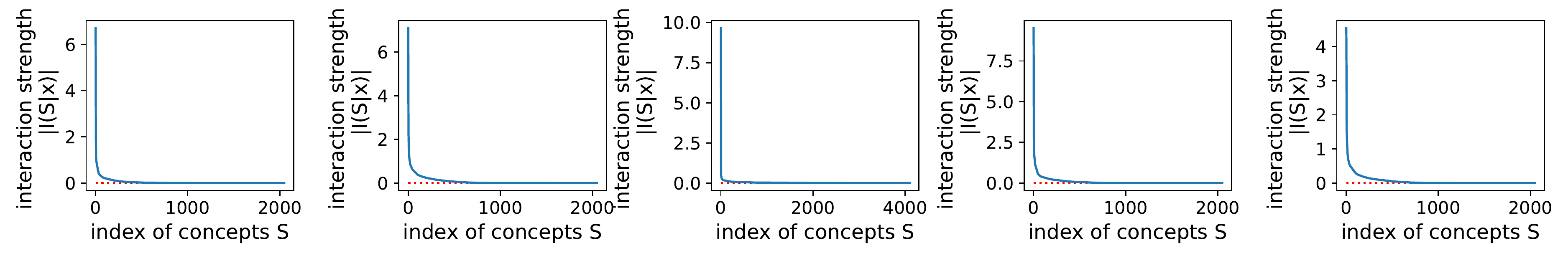}
	\caption{All interactions sorted in descending order of $\lvert I(S|\boldsymbol{x}) \rvert$ values. Each subfigure corresponds to interactions of an input sentence. It shows that only a small number of interactions had salient effects on the LLM's output, and all other interactions had ignorable effects on the LLM's output.}
	\label{fig:sparsity}
\end{figure}

Therefore, in this study, we further examine whether the LLM with much more parameters than traditional DNNs also encodes sparse symbolic concepts. 

We first visualize the distribution of the interaction effects of different concepts encoded by the LLM to examine the sparsity of concepts. In addition, we also examine the claim in Remark~\ref{re:sparsity} that we can use those sparse concepts to well estimate the LLM's outputs on all randomly masked samples. Specifically, we follow \cite{ren2023defining} to extract a few concepts encoded by the LLM. Then, given a randomly masked sentence $\boldsymbol{x}_S$, $S\subseteq N$,
we compare the real output $v^{\text{real}}(\boldsymbol{x}_S)$ and the output $v^{\text{approx}}(\boldsymbol{x}_S)$ approximated by the extracted concepts in Remark~\ref{re:sparsity}. If the extracted concepts well match the LLM's outputs on all masked sentences, then we can consider that the inference score of the LLM can be faithfully explained by symbolic concepts.

To this end, we conducted experiments on the OPT-1.3b model \cite{zhang2022opt} as the target LLM, which contianed 1.3 billion parameters.
We used the LLM to generate a set of input sentences as to construct the testing sentence set, which covered factual knowledge in physics, medicine, art, machine learning, and etc. 
Then, for each testing sentence, we analyzed the probability of the LLM generating the $(n+1)$-th word $y^{\text{truth}}$, when the LLM took the sentence $\boldsymbol{x}$ with the already generated $n$ words as the input. Specifically, we used the method in \cite{li2023does} to quantify the interaction $I(S|\boldsymbol{x})$ as the effect of the concept $S\subseteq N$ on the generation of the target (optimal) word  $y^{\text{truth}}$. Figure~\ref{fig:sparsity} shows all interaction effects of the LLM's inference on different input sentences. It shows that there were only a few salient interactions (\emph{i.e.}, concepts) extracted from an input sentence, and all other interactions had ignorable effects on the LLM's output.

\begin{figure}[tbp]
	\centering
	\includegraphics[width=0.85\linewidth]{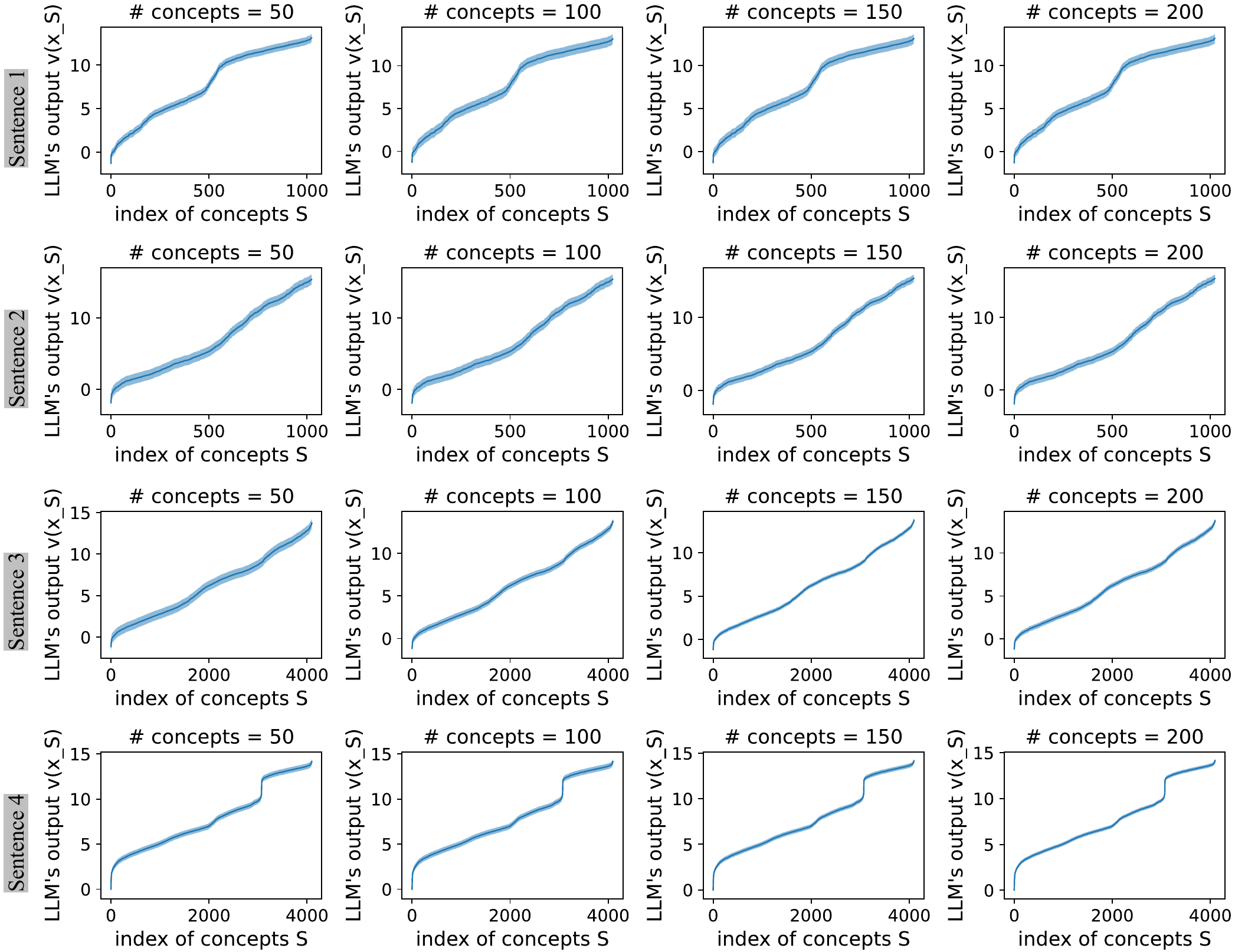}
	\caption{Model output on different masked samples (sorted in an ascending order). The shade area represents the root mean squared error (RMSE) of using the top-ranked concepts to match the real model output. The deviation usually decreased when we used more concepts to match the model output.}
	\label{fig:mimic}
\end{figure}

Besides, we used the method in \cite{ren2023defining,li2023does} to generate the top-ranked 50, 100, 150, and 200 interactions, respectively, as four different sets of concepts to explain the LLM. Then, given each sentence $\boldsymbol{x}$, we constructed a testing set with all the $2^n$ masked sentences $\boldsymbol{x}_S$ by sampling different sets of words $S$.

To verify whether the extracted concepts matched well with the real output $v^{\text{real}}(\boldsymbol{x}_S)$, we measured the standard deviation of approximation errors. Let $\boldsymbol{v}^{\text{real}}=[v^{\text{real}}(\boldsymbol{x}_{S_1}),v^{\text{real}}(\boldsymbol{x}_{S_2}),\ldots,v^{\text{real}}(\boldsymbol{x}_{S_{2^n}})]^{\top}\in\mathbb{R}^{2^n}$ denote the vector of the LLM's outputs on all $2^n$ masked sentences, which were sorted in ascending order. \emph{i.e.}, $v^{\text{real}}(\boldsymbol{x}_{S_1})\le v^{\text{real}}(\boldsymbol{x}_{S_2})\le \cdots \le v^{\text{real}}(\boldsymbol{x}_{S_{2^n}})$. Then, let $\boldsymbol{v}^{\text{approx}}=[v^{\text{approx}}(\boldsymbol{x}_{S_1}),v^{\text{approx}}(\boldsymbol{x}_{S_2}),\ldots,v^{\text{approx}}(\boldsymbol{x}_{S_{2^n}})]^{\top}\in\mathbb{R}^{2^n}$ denote the vector of the outputs approximated by the extracted concepts based on Remark~\ref{re:sparsity}. In this way, $v^{\text{real}}(\boldsymbol{x}_{S_i})-v^{\text{approx}}(\boldsymbol{x}_{S_i})$ represented the matching error of the $i$-th masked sentence. For each model output $v^{\text{real}}(\boldsymbol{x}_{S_i})$, we computed the root mean squared error (RMSE) of the matching errors of the neighboring 50 sentences $\textit{error}_i=\sqrt{ \frac{1}{2t+1}\sum_{j=i-t}^{i+t}(\Delta v_i - \mu_i)^2 }$ to evaluate the matching quality.

Figure~\ref{fig:mimic} reports the real outputs $v^{\text{real}}(\boldsymbol{x}_S)$ and the corresponding RMSE values on all masked sentences. It shows that the approximated outputs matched well with the real outputs on different masked sentences. This proved that the output of the OPT-1.3b model could be explained as sparse symbolic concepts.

$\bullet$ \textbf{Experiment 2: transferability of symbolic concepts encoded by the LLM.}
Besides the sparsity of symbolic concepts, we also analyze the transferability of concepts, which is considered as another important property of concepts.
If a symbolic concept $S$ extracted from a sentence also has a significant effect on the LLM's outputs on other sentences, then the symbolic concept is considered transferable.

Therefore, we conducted experiments to examine the transferability of concepts on the OPT-1.3b model.
We first collected different sets sentences generated by the LLM. All sentences in the same set contained the same word $y^{\text{truth}}$. We aimed to examine whether all words $\boldsymbol{x}^{(i)}$ before $y^{\text{truth}}$ in the $i$-th sentence shared the similar logic. For example, the sentence $\boldsymbol{x}^{(1)}=$ ``\emph{Diabetes is a disease that affects the body's ability to use}'' and the sentence $\boldsymbol{x}^{(2)}=$ ``\emph{Diabetes is a condition that impairs the body's ability to use}'' were supposed to share the same logic to push the LLM to generate the word $y^{\text{truth}}=$ ``\emph{glucoses}.''

In this way, we defined the transferability of concepts between sentences $\boldsymbol{x}^{(1)}$ and $\boldsymbol{x}^{(2)}$ as the similarity between the distribution of concepts extracted from these two sentences. Let $N^{(1)}$ and $N^{(2)}$ denote the sets of all words in $\boldsymbol{x}^{(1)}$ and $\boldsymbol{x}^{(2)}$, respectively. Then, $N^{\text{same}}= N^{(1)} \cap N^{(2)}$ denotes the set of words shared by the two sentences.
Then, we followed Li et al. \cite{li2023does} to extract the top-ranked $M$ salient interactions from the first sentence $\boldsymbol{x}^{(1)}$, \emph{i.e.}, the set of $M$ interactions with the largest $\lvert I(S|\boldsymbol{x})\rvert$ values, denoted by $\Omega_{\text{salient}}^{(1)}$.
Similarly, $\Omega_{\text{salient}}^{(2)}$ denotes the set of top-ranked $M$ salient concepts extracted from the second sentence $\boldsymbol{x}^{(2)}$. To compute the similarity between concepts in $\Omega_{\text{salient}}^{(1)}$ and $\Omega_{\text{salient}}^{(2)}$, we constructed the vector $I^{(1)}=[I^{(1)}_{S_1},I^{(1)}_{S_2},\ldots,I^{(1)}_{S_d},\tilde{I}^{(1)}_{S_1},\tilde{I}^{(1)}_{S_2},\ldots,\tilde{I}^{(1)}_{S_d}]\in\mathbb{R}^{2d}$ and the vector $I^{(2)}=[I^{(2)}_{S_1},I^{(2)}_{S_2},\ldots,I^{(2)}_{S_d},\tilde{I}^{(2)}_{S_1},\tilde{I}^{(2)}_{S_2},\ldots,\tilde{I}^{(2)}_{S_d}]\in\mathbb{R}^{2d}$, where $S_1,S_2,\ldots,S_d$ represented all the $d=2^{\vert N^{\textrm{same}}\vert}$ interactions between words in $N^{\text{same}}$. If $S_t \in \Omega_{\text{salient}}^{(1)}$, we set $I^{(1)}_{S_t} = \max(I(S_t | \boldsymbol{x}),0)$ and $\tilde{I}^{(1)}_{S_t} = -\min(I(S_t | \boldsymbol{x}),0)$; otherwise, we set $I^{(1)}_{S_t} = \tilde{I}^{(1)}_{S_t} = 0$. Then, the similarity of the concept distribution between $\boldsymbol{x}^{(1)}$ and $\boldsymbol{x}^{(2)}$ was given as the Jaccard similarity between $I^{(1)}$ and $I^{(2)}$, \emph{i.e.}, $\text{sim}(I^{(1)},I^{(2)}) = \frac{\lVert \min (I^{(1)},I^{(2)}) \rVert_{1}}{\lVert \max(I^{(1)},I^{(2)}) \rVert_{1}}$. Thus, a high similarity $\text{sim}(I^{(1)},I^{(2)})$ indicated that most concepts shared by sentences $\boldsymbol{x}^{(1)}$ and $\boldsymbol{x}^{(2)}$ had considerable transferability.

\begin{table}[t]
	\caption{Similarity between the distribution of concepts extracted from different sentences. $M$ denotes the number of concepts extracted from each sentence.}
	\vspace{-5pt}
	\label{tab:exp2}
	\centering
	\resizebox{0.8\linewidth}{!}
	{
		\begin{tabular}{lcccccc}
			\toprule
			& M = 5	& M = 10 & M = 15 & M = 20 &   M = 25 & M = 30 \\
			\midrule
			average $\text{sim}(I^{(1)},I^{(2)})$ & 0.565 $\pm$ {\small 0.202} & 0.528 $\pm$ {\small 0.205} & 0.476 $\pm$ {\small 0.184} & 0.454 $\pm$ {\small 0.183} & 0.433 $\pm$ {\small 0.179} & 0.416 $\pm$ {\small 0.173} \\
			\bottomrule
		\end{tabular}
	}
\end{table}

\begin{table}[t]
	\caption{Symbolic concepts extracted from input sentences in the LLM.}
	\vspace{-5pt}
	\label{tab:exp3}
	\centering
	\resizebox{\linewidth}{!}
	{
		\begin{tabular}{cc|cc}
			\toprule
			\multicolumn{2}{l|}{Sentence 1: The human brain contains approximately 100 billion neurons,}	& \multicolumn{2}{l}{Sentence 2: The force required to accelerate an object is directly proportional}   \\
			\multicolumn{2}{l|}{which communicate with each other through electrical and}	& \multicolumn{2}{l}{to its mass and acceleration, as described by Newton's laws of}   \\
			\multicolumn{2}{l|}{Predicted word: chemical}	& \multicolumn{2}{l}{Predicted word: motion}   \\
			\midrule
			Concept $S$ & Inference effect $I(S|\boldsymbol{x})$ & Concept $S$ & Inference effect $I(S|\boldsymbol{x})$ \\
			\midrule
			\{\emph{electrical, and}\} & 4.82 &\{\emph{laws}\} & 5.99 \\
			\{\emph{electrical}\} & 3.46 &\{\emph{Newton's}\} & 3.73 \\
			\{\emph{through, electrical, and}\} & -2.37 & \{\emph{Newton's, laws}\} & 1.10\\
			\{\emph{brain}\} & 1.11  & \{\emph{acceleration}\} & 0.86\\
			\{\emph{brain, electrical, and}\} & 1.08  & \{\emph{force}\} & 0.61\\
			\midrule
			\midrule
			\multicolumn{2}{l|}{Sentence 3: The hepatitis B virus is a highly infectious blood-borne virus}	& \multicolumn{2}{l}{Sentence 4: Newton's laws of motion state that an object will remain at rest}   \\
			\multicolumn{2}{l|}{that can cause serious}	& \multicolumn{2}{l}{ or in uniform motion in a straight}   \\
			\multicolumn{2}{l|}{Predicted word: liver}	& \multicolumn{2}{l}{Predicted word: line}   \\
			\midrule
			Concept $S$ & Inference effect $I(S|\boldsymbol{x})$ & Concept $S$ & Inference effect $I(S|\boldsymbol{x})$ \\
			\midrule
			\{\emph{serious}\} & 4.54 & \{\emph{straight}\} & 9.65\\
			\{\emph{hepatitis B}\} & 3.99 & \{\emph{laws, straight}\} & 0.62 \\
			\{\emph{hepatitis B, virus, highly, infectious, blood, cause}\} & 1.49 & \{\emph{Newton's, straight}\} & 0.48 \\
			\{\emph{cause}\} & 1.32 & \{\emph{object, straight}\} &0.37 \\
			\{\emph{hepatitis B, serious}\} &  1.13 & \{\emph{object, remain, straight}\} & -0.35 \\
			\bottomrule
		\end{tabular}
	}
\end{table}

Table~\ref{tab:exp2} reports the average $\text{sim}(I^{(1)},I^{(2)})$ over different pairs of sentences $(\boldsymbol{x}^{(1)},\boldsymbol{x}^{(2)})$. Results show that top-ranked salient concepts usually exihibited relatively significant transferability.

$\bullet$ \textbf{Experiment 3: discrimination power of symbolic concepts encoded by the LLM.}
Ideally, a faithful symbolic concept is supposed to push the LLM to generate the sentence that conforms to the factual knowledge. Therefore, Table~\ref{tab:exp3} shows concepts that had the most significant effects on the LLM's inference. Results show that most concepts were consistent with human cognition.

$\bullet$ \textbf{Experiment 4: examining the exact reasons accountable for the LLM's prediction errors.} 
Examining the exact reasons why an LLM generated a sentence against the fact is the basis for debugging the LLM. As proved in Experiment 1, the inference score of an LLM could be disentangled into a few symbolic concepts. Therefore, in this paper, we extracted the exact reasons accountable for errors in the generated sentences.
For example, as shown in Table~\ref{tab:exp4}, the LLM generated the sentence ``\emph{Physicist Isaac Newton was born in 1642 in the village of Newton},'' where ``\emph{in the village of Newton}'' ran counter to the fact of ``in Woolsthorpe-by-Colsterworth.'' Then, we analyzed the reason why the LLM generated the $(n+1)$-th word that ran counter to the fact, denoted by $y^{\text{counter}}$. We extracted all concepts that had signifcant-yet-positive effects on the generation of $y^{\text{counter}}$, \emph{i.e.}, $S\in\Omega_{\text{salient}},I(S|\boldsymbol{x})>0$, as accountable reasons for the error. Table~\ref{tab:exp4} shows two examples of the OPT-1.3b model's prediction errors. In the first example, the concept of \{\emph{Newton}\} (a person's name) made the largest effect on the LLM's prediction error of ``Newton'' (a place name), but the correct answer was ``Woolsthorpe-by-Colsterworth.'' It might be because the OPT-1.3b model had not studied biographies, so it could not answer such questions. In the second example, the concept of \{\emph{Drake}\} made the largest effect on the OPT-1.3b model's prediction error of ``Drake,'' but the correct answer was ``Mike.'' We found that it was the line break symbol ``$\backslash$n'' that affected the OPT-1.3b model's encoding of the word ``\emph{Mike}.'' Without the line break symbol ``$\backslash$n,'' the OPT-1.3b model would encode the concept of \{\emph{Mike}\}, and generate the word ``\emph{Mike}.''

\begin{table}[t]
	\caption{Symbolic concepts accountable for the LLM's prediction errors.}
	\label{tab:exp4}
	\centering
	\resizebox{\linewidth}{!}
	{
		\begin{tabular}{cc|cc}
			\toprule
			\multicolumn{2}{l|}{Sentence 1: Physicist Isaac Newton was born in the village of}	& \multicolumn{2}{l}{Sentence 2: $\backslash$n Mike's mum had 3 kids; 2 of them are Luis and Drake.}   \\
			\multicolumn{2}{l|}{Predicted word: Newton}	& \multicolumn{2}{l}{The name of remain kid is}   \\
			&	& \multicolumn{2}{l}{Predicted word: Drake}   \\
			\midrule
			Concept $S$ & Inference effect $I(S|\boldsymbol{x})$ & Concept $S$ & Inference effect $I(S|\boldsymbol{x})$ \\
			\midrule
			\{\emph{Newton}\} & 6.26 & \{\emph{Drake}\} & 6.94 \\
			\{\emph{village, of}\} & 1.57 &  \{\emph{name}\} & 1.15\\
			\{\emph{Physicist}\} & 1.29 &  \{\emph{kids, Drake}\} & 0.64\\
			\{\emph{of}\} & 1.18 &  \{\emph{are, Luis}\} & 0.58\\
			\{\emph{village}\} & 1.02 &  \{\emph{mum, Drake}\} & 0.42 \\
			\bottomrule
		\end{tabular}
	}
\end{table}

\section{Conclusion}

In this paper, we have analyzed the symbolic concepts encoded by an LLM for dialogue. Specifically, we have empirically verified that the inference score of an LLM can be disentangled into a small number of concepts. Those symbolic concepts usually exhibit high transferability and exhibit certain
discrimination power. More crucially, we have also used those symbolic concepts to explain the prediction errors of an LLM.

{
\bibliographystyle{plain}
\bibliography{ref}
}

\end{document}